\documentclass[10pt,twocolumn,letterpaper]{article}

\usepackage[pagenumbers]{wacv} 

\usepackage{amsmath}
\usepackage{amssymb}

\usepackage{algorithm}
\usepackage{algpseudocode}

\usepackage[table]{xcolor}
\usepackage{colortbl}

\usepackage{pgfplots}
\pgfplotsset{compat=1.18}

\definecolor{hlgreen}{RGB}{220, 245, 225}

\usepackage{multirow}

\definecolor{wacvblue}{rgb}{0.21,0.49,0.74}
\usepackage[pagebackref,breaklinks,colorlinks,allcolors=wacvblue]{hyperref}

\title{SARFA: Segment Anything with Radiomic Feature Alignment}

\author{
    \begin{tabular}{@{\hspace{0.5in}}c@{\hspace{1.6in}}c@{\hspace{0.5in}}}
    Tyler Ward & Abdullah Imran\\
    {\tt\small tyler.ward@uky.edu} & {\tt\small aimran@uky.edu}
    \end{tabular}\\[2em]
    Computer Science Department\\
    University of Kentucky, Lexington, KY, USA
}

\begin{document}

\maketitle

\begin{abstract}
The Segment Anything Model (SAM) has demonstrated strong generalizability across a variety of segmentation tasks. However, SAM often struggles in situations where the target to be segmented is ambiguous. This poses a problem in medical imaging, where accurate delineation of targets such as tumors is vital, but even expert radiologists can disagree on the appropriate boundary for a target. Addressing this, we propose \textbf{SARFA} (\textbf{S}egment \textbf{A}nything with \textbf{R}adiomic \textbf{F}eature \textbf{A}lignment), a novel framework for improved medical image segmentation. Via probabilistic prompting, SARFA generates a diverse set of plausible masks for each input image and optimizes them with a radiomics-driven training objective based on Fréchet Radiomic Distance (FRD) and Direct Preference Optimization (DPO). By minimizing the FRD between masked predicted and ground truth regions within each image, SARFA encourages segmentation outputs whose anatomical and textural characteristics align with clinically meaningful ground truth representations, without relying solely on pixel-level overlap. Evaluated on computed tomography (CT) and magnetic resonance imaging (MRI) benchmarks, SARFA outperforms existing ambiguous segmentation methods, demonstrating the effectiveness of radiomic feature alignment and DPO-style candidate mask ranking as a training objective. Our code is available at \url{https://github.com/tbwa233/SARFA}.
\end{abstract}

\section{Introduction}
\label{sec:intro}

In medical image analysis, segmentation is commonly used to isolate regions of interest from the background, which helps to identify lesions as well as their sizes, locations, and relationships with surrounding tissues~\cite{gao2025medical}. Despite great reductions in the time it takes to perform medical image segmentation due to advances in deep learning, training models to recognize difficult targets can still be challenging due to the limited availability of expert annotations and the inherent ambiguity that exists in many medical imaging tasks~\cite{fasihi2016overview}. 

Ambiguity in medical images can arise from numerous sources, which we lump into three different categories: boundary ambiguity, inter-rater ambiguity, and aleatoric. Based on established literature~\cite{xi2025uncertainty}, we define boundary ambiguity as occurring when lesion margins are difficult to distinguish from surrounding tissue; inter-rater ambiguity as occurring when multiple experts provide differing yet clinically accepted annotations; and aleatoric uncertainty as occurring when the image itself contains insufficient information to uniquely determine the target boundary. As a result of these potential sources of ambiguity, there may exist multiple plausible segmentations for a single image~\cite{rahman2023ambiguous}. This poses a problem for conventionally trained segmentation models, as these models are largely deterministic in the sense that they produce just one output mask~\cite{zhou2025udndsnet}. As a result, there has been much research into the development of probabilistic, ambiguity-aware segmentation models for medical imaging that explicitly model distributions of plausible annotations~\cite{baumgartner2019phiseg, gao2023modeling, huang2024p2sam, kassapis2021calibrated, kohl2018probabilistic, kohl2019hierarchical, monteiro2020stochastic, rahman2023ambiguous, ward2026probabilistic, zhang2022pixelseg, zhang2023customized}.

Recently, advances in large-scale data training have enabled foundation models like the Segment Anything Model (SAM)~\cite{kirillov2023segment} to achieve strong generalization across a wide range of segmentation tasks. However, applying such models in medical imaging is difficult due to the substantially different characteristics of medical images vs. natural images, including weak differentiation between targets as a result of subtle contrast changes~\cite{ma2024segment}. There remains a need for methods leveraging the representational power of foundation models while simultaneously accounting for the ambiguity inherent to medical image segmentation.

\begin{figure*}[t]
    \centering
    \includegraphics[width=\linewidth]{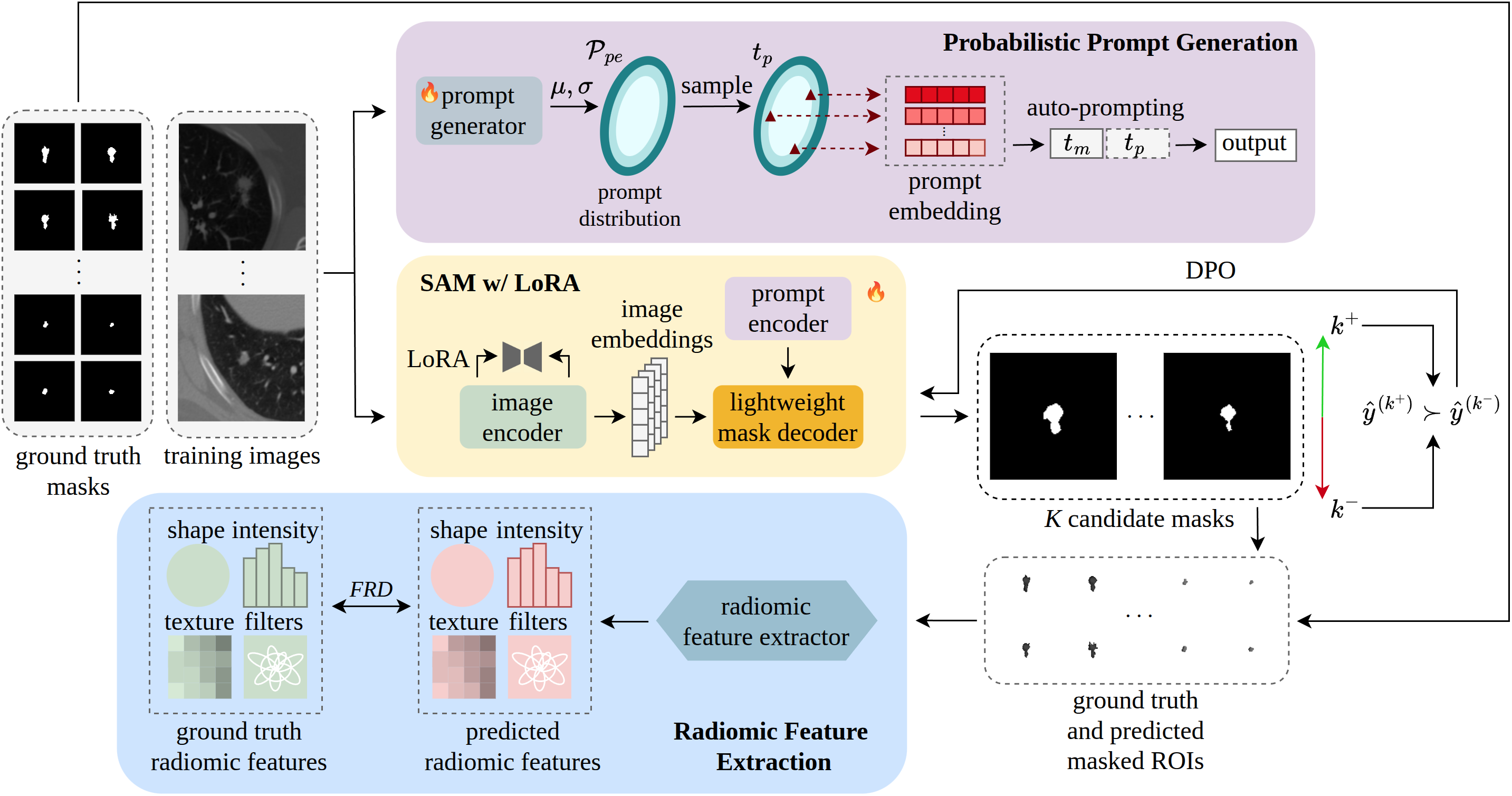}
    \caption{Overview of the proposed SARFA framework. A probabilistic prompt generator produces diverse prompt embeddings that enable LoRA-adapted SAM to generate $K$ candidate segmentation masks for each input image. Masked regions from both ground truth and predicted candidates are passed through a radiomic feature extraction pipeline, and their similarity is quantified using FRD. The radiomic distances are used to rank masks, defining preferred and rejected candidates for our DPO loss, which encourages anatomically and texturally consistent medical image segmentation.}
    \label{fig:saarch}
\end{figure*}

A largely unexplored opportunity lies in the use of radiomic information as a supervisory signal during model training. Radiomic features are quantitative descriptors that characterize image intensity distributions, texture patterns, morphological properties, and higher-order image characteristics that may not be readily apparent to human observers~\cite{mayerhoefer2020introduction}. Because these features capture information beyond simple spatial overlap, they offer a potentially valuable representation for comparing alternative segmentation hypotheses. Specifically, we \textit{hypothesize} that training a model to minimize differences in ground truth and predicted radiomic feature distributions can lead to more accurate segmentation compared to methods that rely solely on maximizing pixel-level overlap.
To this end, we introduce \textbf{SARFA} (\textbf{S}egment \textbf{A}nything with \textbf{R}adiomic \textbf{F}eature \textbf{A}lignment), whose architecture is seen in Fig.~\ref{fig:saarch}. Our specific contributions can be summarized as follows:

\begin{itemize}
    \item We address annotation unavailability and ambiguity by leveraging SAM's multimask decoding mechanism to generate multiple plausible segmentation candidates;
    \item Integration of radiomics-based supervision into the segmentation pipeline using PyRadiomics~\cite{van2017computational}-based extraction and calculation of the Fréchet Radiomic Distance (FRD)~\cite{konz2024fr};
    \item Optimization of SAM predictions using a novel mask ranking system and direct preference optimization (DPO)~\cite{rafailov2023direct}-style loss function; 
    \item Validation on both computed tomography (CT) scans and magnetic resonance imaging (MRI) demonstrates the superiority of SARFA over existing state-of-the-art (SOTA) baselines. This also indicates that radiomic feature alignment through the minimization of FRD and utilization of DPO-style candidate mask ranking can be an effective training objective.
\end{itemize}

\section{Related Work}
\label{sec:related_work}

\subsection{Ambiguous Medical Image Segmentation}

There is an inherent level of ambiguity that exists within medical image segmentation problems that is not adequately addressed by deterministic segmentation methods. Because of this, several research pathways have been developed over the years in an effort to solve this problem. Early work was dominated by conditional variational autoencoder (cVAE)-based approaches, with Probabilistic U-Net~\cite{kohl2018probabilistic} first demonstrating that multiple coherent segmentation hypotheses could be generated from a learned latent space, while later methods such as the Hierarchical Probabilistic U-Net (HPU-Net)~\cite{kohl2019hierarchical} and PHiSeg~\cite{baumgartner2019phiseg} introduced hierarchical latent representations to better capture ambiguity across spatial scales.

Subsequent research explored alternative uncertainty modeling strategies, including correlated probabilistic representations~\cite{monteiro2020stochastic}, adversarial refinement~\cite{kassapis2021calibrated}, autoregressive generation~\cite{zhang2022pixelseg}, diffusion models~\cite{rahman2023ambiguous}, and mixtures of stochastic experts~\cite{gao2023modeling}. Collectively, these methods improved the diversity, calibration, and realism of generated segmentation hypotheses while highlighting the importance of modeling ambiguity as a structured distribution rather than as independent pixel-wise uncertainty.

More recent works have explored the capabilities of foundation models. SAMed~\cite{zhang2023customized} demonstrated that SAM can be effectively adapted for medical image segmentation, while P$^2$SAM~\cite{huang2024p2sam} and Probabilistic SAM~\cite{ward2026probabilistic} extend SAM to ambiguity-aware segmentation by introducing probabilistic prompting and latent-variable modeling, respectively.

\subsection{Prompting the Segment Anything Model}

A defining characteristic of SAM is its reliance on prompts. While this enables great flexibility, it poses a problem for medical imaging tasks, where accurate prompting would require expert knowledge. As a result, a growing body of research has focused on reducing or eliminating the need for manual prompting. One such approach, employed by techniques such as AutoSAM~\cite{shaharabany2023autosam}, MaskSAM~\cite{xie2025masksam}, SAM-SP~\cite{zhou2024sam}, and Sam2Rad~\cite{wahd2025sam2rad}, relies on the learning or prediction of prompts directly from image representations, greatly enhancing the autonomy of SAM-based segmentation frameworks. Other methods rely on the automatic generation of prompts from auxiliary task information~\cite{ward2025annotation, ward2025autoadaptive}.

From the last method, the question naturally arises: \textit{how to best utilize such preference data within a SAM-based segmentation pipeline}. This is largely under-explored in the literature, although there have been some efforts. Konwer \textit{et al.}~\cite{konwer2025enhancing} combined automated prompting with DPO, demonstrating that segmentation models can be improved through ranked candidate outputs rather than explicit reward functions. While effective, this approach relied on a complex pipeline comprising multiple external networks and components, leaving room for efficiency improvements.

\subsection{Radiomics Information as a Supervisory Signal}

Recognizing the limitations of segmentation methods that rely solely on pixel-level overlap for assessment of segmentation quality, there exists a growing body of work that focuses on leveraging radiomic features within deep learning pipelines. One of the prominent methodologies towards this end is multi-task/dual-branch architectures. For example, Hambarde \textit{et al.}~\cite{hambarde2020prostate} map semantic segmentations alongside statistical radiomic representations to ensure boundaries enclose distinct microstructures. Similar joint optimization approaches fuse radiomic features with deep semantic embeddings to supervise complex clinical downstream prognostic tasks, such as glioma sub-volume segmentation and breast cancer neoadjuvant chemotherapy response prediction~\cite{chen2023radiomics, hao2023predicting}. Such approaches force models to preserve clinical shape, contrast, and phenotypic tumor signatures, yielding superior classification calibration over networks optimized purely on standard imaging features~\cite{tang2023artificial, lekkas2025deep}.

Alternatively, radiomics can be applied directly via auxiliary regularization and alignment loss functions during backpropagation to enhance robustness. Instead of training separate prediction branches, these networks introduce structural loss functions that penalize discrepancies between the ground-truth and predicted regions' sub-visual gray-level textures. For example, Yang \textit{et al.}~\cite{yang2022deep} implement a radiomics-based support vector machine (SVM) alongside a Siamese network with contrastive and cross-entropy losses to evaluate the geometry and texture of 3D volumes, effectively isolating true lesions from false positives. Similarly, Bhattacharya \textit{et al.}~\cite{bhattacharya2022gazeradar} introduced a novel radiomics-visual attention loss (RVAL) to enforce consistency across a shared, domain-invariant radiomic metric space. This effectively calculates the distance between predicted and reference feature distributions, regularizing spatial focus and making the network less sensitive to scanner-specific variations.

This quantitative feature encoding and structural relationship modeling has seen expanding utility across modern architectures, including graph neural networks for lung CT profiling~\cite{faizi2025graph}, myocardial infarction mapping on cine-CMR~\cite{xu2025integrating}, adaptable Gabor and Laplacian of Gaussian filtered transformers for abdominal organ segmentation~\cite{zarch2025glog}, pelvic adnexal mass ultrasound classification~\cite{barcroft2024machine}, global-to-voxel parametric map injections for pancreatic lesions~\cite{deng2026global}, and adversarial networks for volumetric lesion generation~\cite{li2024generative}. However, existing methods typically apply these signatures post hoc or within rigid, deterministic architectures. To the best of our knowledge, SARFA is the first framework to leverage radiomics-driven supervision within a foundation model architecture like SAM.

\section{Methods}

Let $\mathcal{D} = \{ (x_i, y_i) \}_{i=1}^{N}$ denote a dataset of medical images, $x_i \in \mathbb{R}^{H \times W}$, and corresponding ground-truth segmentation masks, $y_i \in \{0,1\}^{H \times W}$. Our goal is to learn a segmentation model, $G_{\theta}$, parameterized by $\theta$, such that $G_{\theta}(x_i) \rightarrow \hat{y}_i$, where $\hat{y}_i$ approximates $y_i$ while preserving clinically relevant radiomic characteristics.

\subsection{Probabilistic Prompt Generation}

Recent work in fine-tuning SAM for medical image segmentation, particularly in low-label or ambiguously labeled settings, has explored various methods to optimize prompt generation. One such method lies in the generation of multiple segmentation candidates by thresholding mask probabilities and subsequently applying preference optimization over these proposals~\cite{konwer2025enhancing}. While effective, this strategy constructs candidates post hoc from a single deterministic mask distribution. An alternative method relies on the introduction of a prior probabilistic space for prompts~\cite{huang2024p2sam}. This allows for the generation of ``one-to-many'' segmentation mappings by sampling from a learned prompt distribution. In this work, we modify this probabilistic prompting strategy.

The SAM architecture consists of an image decoder, $\text{Enc}_I$, a prompt encoder, $\text{Enc}_P$, and a mask decoder, $\text{Dec}_M$. From $\text{Enc}_I$, we obtain:

\begin{equation}
    F_I = \text{Enc}_I(x),
\end{equation}

\noindent where $F_I \in \mathbb{R}^{h \times w \times c}$ is the image embedding. Simultaneously, $\text{Dec}_M$ produces $K$ mask tokens, 

\begin{equation}
\{ \hat{y}^{(k)} \}_{k=1}^{K} \text{Dec}_M(F_I, T_M), 
\end{equation}

\noindent where $T_M$ denotes the set of learned tokens. 

Following the probabilistic prompting strategy of $\text{P}^2\text{SAM}$~\cite{huang2024p2sam}, we interpret these multiple masks as samples from an implicit segmentation distribution, 

\begin{equation}
    \tilde{Y} \sim P_{\theta}(\tilde{Y} \mid x),
\end{equation}

\noindent where the distribution $P_{\theta}$ is induced by the interaction between $F_I$ and $T_M$. Unlike $\text{P}^2\text{SAM}$, which explicitly models a Gaussian prompt embedding distribution and samples prompt embeddings, our approach leverages SAM’s built-in multimask decoding to approximate this sampling process,

\begin{equation}
    \mathbb{E}_{\tilde{Y} \sim P_{\theta}(\cdot \mid x)} \left[ \tilde{Y} \right] \approx \frac{1}{K} \sum_{k=1}^{K} \hat{y}^{(k)}.
\end{equation}

From here, we introduce a lightweight learnable mask weighting mechanism that operates directly over SAM’s multimask outputs. This can be expressed mathematically by allowing $w_k \in \mathbb{R}$ to denote learnable weights such that: 

\begin{equation}
    \sum_{k=1}^{K} w_k = 1.
\end{equation}

\noindent The final segmentation prediction in this case can be expressed as, 

\begin{equation}
    \tilde{y} = \sum_{k=1}^{K} w_k \hat{y}^{(k)}.
\end{equation}

\noindent Such a formulation preserves SAM’s intrinsic ambiguity while allowing the model to learn task-specific scale and structural preferences without explicitly modeling a prompt distribution.

\subsection{Radiomic Feature Extraction}

FRD~\cite{konz2024fr} is a task-independent metric for comparing medical image distributions using radiomic features instead of learned natural-image embeddings. In this work, we extract radiomic features from masked regions of the input images, $x_m$, where $x_m$ is formed from the element-wise multiplication of $x$ and $m$. Here, $m$ can represent either the ground truth mask or one of the predicted candidate masks. Extracting the features from just the region of the image overlapped by the ground truth/predicted mask ensures that the features correspond specifically to the predicted anatomical structure. The extraction process is handled by a standard PyRadiomics~\cite{van2017computational} pipeline, and extracted features include first-order statistics, texture features, Shape2D features, and wavelet-transformed feature maps. After extraction, we are left with the following feature vector: 

\begin{equation}
    \Phi(x,m) \in \mathbb{R}^d.
\end{equation}

Following the normalization strategy used by Konz \textit{et al.}~\cite{konz2024fr}, we compute dataset-level statistics over the ground truth masks, then \textit{z-score} normalize each feature vector, 

\begin{equation}
    \bar{z}=\frac{\Phi(x,m)-\mu}{\sigma}.
\end{equation}

As our model produces $K$ candidate masks for each image $x$, we computed these normalized radiomic features for the ground truth and each candidate mask:

\begin{equation}
    z_{\text{gt}} = \Phi(x, y),
    \qquad
    z^{(k)} = \Phi(x, \hat{y}^{(k)}).
\end{equation}

After normalization, we measure the distance between the radiomic features of each candidate mask and the ground truth masks' radiomic features. We do this by calculating the mean squared distance between them in the feature space: 

\begin{equation}
    d^{(k)} = \left\| \bar{z}^{(k)} - \bar{z}_{\text{gt}} \right\|_2^2.
\end{equation}

Once the radiomic distances are known, we select a ``preferred'' and ``rejected'' mask, like so: 

\begin{equation}
    k^+ = \arg\min_{k} d^{(k)},
    \qquad
    k^- = \arg\max_{k} d^{(k)}.  
\label{eq:preferred_vs_rejected_masks}
\end{equation}

\noindent where $k^+$, the preferred mask, is the candidate mask that has the lowest radiomic distance to the ground truth radiomic features, and the rejected mask, $k^-$, has the highest. Both $k^+$ and $k^-$ are used in the calculation of our DPO loss, discussed in the next section.

In addition to per-sample mask ranking, we evaluate model performance at the epoch level using FRD. Given radiomic feature sets $D_{gt}$ and $D_{pred}$ for the ground truth and predicted masked image regions, we compute their empirical means and covariances:

\begin{equation}
    (\mu_{\text{gt}}, \Sigma_{\text{gt}}),
    \qquad
    (\mu_{\text{pred}}, \Sigma_{\text{pred}}).
\end{equation}

\noindent The FRD is then:

\begin{equation}
\begin{aligned}
    \mathrm{FRD}(\mathcal{D}_{\text{gt}}, \mathcal{D}_{\text{pred}})
    =
    \left\| \mu_{\text{gt}} - \mu_{\text{pred}} \right\|_2^2
    \\+
    \mathrm{Tr}\!\left(
    \Sigma_{\text{gt}} + \Sigma_{\text{pred}}
    - 2\left( \Sigma_{\text{gt}} \Sigma_{\text{pred}} \right)^{1/2}
    \right),
\end{aligned}
\end{equation}

\noindent which corresponds to the 2-Wasserstein distance between Gaussian approximations of the radiomic feature distributions. In our approach, this epoch-level FRD calculation is used to determine which model checkpoint to save, with the checkpoint with the lowest FRD being the one saved and used for evaluation.

\subsection{Direct Preference Optimization}

DPO is an alternative to reinforcement learning from human feedback (RLHF) that eliminates the need for an explicit reward model by directly optimizing a policy from pairwise preference data~\cite{rafailov2023direct}. Initially proposed for language modeling, recent work has demonstrated the efficacy of DPO when used as a training objective for segmentation models~\cite{konwer2025enhancing}. Here, we utilize DPO to help our proposed SARFA learn to prefer masks that have the lowest radiomic distance. 

Following the calculation of the radiomic distances, we are left with $k^+$ and $k^-$, the masks with the lowest and highest radiomic distances to the ground truth, respectively. This defines a pairwise preference, 

\begin{equation}
    \hat{y}^{(k^{+})} \succ \hat{y}^{(k^{-})}. 
\end{equation}

For our DPO implementation, let $\pi_\theta$ denote the current policy model and $\pi_{\text{ref}}$ denote a frozen reference model initialized from the same weights. In our implementation, the ``policy'' corresponds to the IoU prediction head of SAM, which outputs scores over the $K$ candidate masks. These scores are converted to probabilities using a softmax. So, for a preferred mask $k^+$ and rejected mask $k^-$ we define:

\begin{equation}
    \log p_\theta(k | x) = \log\text{softmax}(s_\theta(x))_k, 
\end{equation}

\noindent where $s_\theta(x)$ are the IoU logits.

The DPO loss for a single image is:

\begin{equation}
\begin{aligned}
\mathcal{L}_{\text{DPO}}
=
- \log \sigma \Big(
\beta \big[
&\big(
\log p_{\theta}(k^{+} \mid x)
-
\log p_{\theta}(k^{-} \mid x)
\big) \\
&-
\big(
\log p_{\text{ref}}(k^{+} \mid x)
-
\log p_{\text{ref}}(k^{-} \mid x)
\big)
\big]
\Big).
\end{aligned}
\end{equation}

\noindent where $\sigma(\cdot)$ is the logistic function and $\beta$ is a temperature parameter controlling preference strength. The full training loss combines supervised segmentation losses with DPO, 

\begin{equation}
    \mathcal{L}_\text{total} = \mathcal{L}_\text{sup} + (\lambda \times \mathcal{L}_\text{DPO}), 
\end{equation}

\noindent where $\mathcal{L}_\text{sup}$ includes cross-entropy, Dice, and Focal losses, and $\lambda$ is a small weight applied to the DPO loss. To ensure stability in the loss calculation, we apply DPO intermittently during training every $T$ steps.

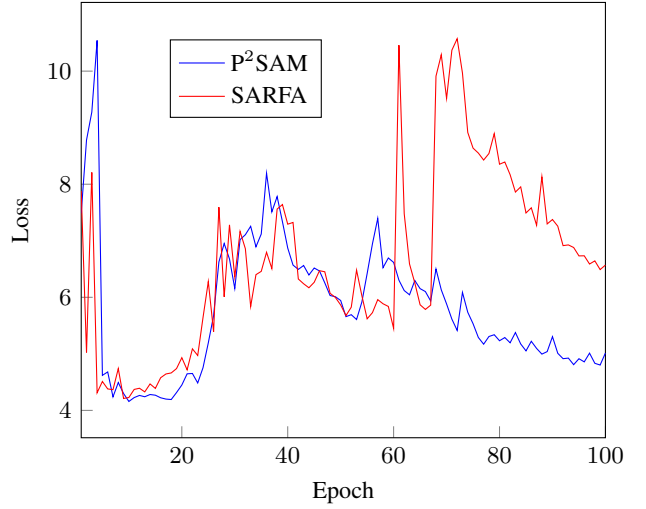
\begin{figure}[t]
\centering
\resizebox{\linewidth}{!}{%
\begin{tikzpicture}
\begin{axis}[
    font=\small,
    xlabel={Epoch},
    ylabel={Loss},
    xmin=1,
    xmax=100,
    enlarge x limits=false,
    grid=none,
    tick pos=left,
    legend style={
        at={(0.314,0.825)},
        anchor=center,
        draw=black,
        fill=white,
        fill opacity=1,
        text opacity=1
    }
]

\addplot[
    blue,
    mark=none
] table[x=epoch,y=train,col sep=space] {losses.dat};
\addlegendentry{P$^2$SAM}

\addplot[
    red,
    mark=none
] table[x=epoch,y=second,col sep=space] {losses.dat};
\addlegendentry{SARFA}

\end{axis}
\end{tikzpicture}%
}
\caption{Progression of loss values for P$^2$SAM and SARFA when trained for 100 epochs on the LIDC-IDRI. Both models exhibit rapid convergence within the first 10 epochs; hence, the selection of this value for the rest of our experiments.}
\label{fig:loss_curves}
\end{figure}

\section{Experimental Evaluation}
\label{sec:implementationdetails}

\subsection{Data}

We validate our proposed SARFA on the Lung Image Database Consortium and Image Database (LIDC-IDRI)~\cite{armato2011lung}, which contains lesion annotations collected from four expert radiologists across 1,018 lung CT scans from 1,010 patients. To demonstrate efficacy across imaging modalities, we also evaluate our model on data from the 2017 Brain Tumor Segmentation (BraTS) Challenge~\cite{menze2014multimodal}. This dataset contains annotations for GD-enhancing tumor, peritumoral edema, and necrotic/non-enhancing tumor across 285 3D MRI images, comprised of 155 slices in four modalities (T1, T1ce, T2, and FLAIR). Following Huang \textit{et al.}'s  method~\cite{huang2024p2sam}, we treat each of the three ground truth masks per slice as a unique ambiguous mask to be passed to SARFA. We follow the train/val/test splits in~\cite{huang2024p2sam}.

\begin{table}[t]
\centering
\caption{Comparison of SOTA ambiguous segmentation models on LIDC. The best-performing method across the majority of metrics is highlighted in green.}
\resizebox{\linewidth}{!}{
\Large
\begin{tabular}{lcccc}
\toprule
Method & GED($\downarrow$) & FRD($\downarrow$) & HM-IoU($\uparrow$) & $D_{\max}(\uparrow)$ \\
\midrule
Probabilistic U-Net~\cite{kohl2018probabilistic} & 0.324 & \textbf{--} & 0.423 & 0.370 \\
HPU-Net~\cite{kohl2019hierarchical} & 0.270 & \textbf{--} & 0.530 & \textbf{--} \\
PHiseg~\cite{baumgartner2019phiseg} & 0.262 & \textbf{--} & 0.595 & \textbf{--}    \\
SSN~\cite{monteiro2020stochastic} & 0.259 & \textbf{--} & 0.555 & \textbf{--} \\
CAR~\cite{kassapis2021calibrated} & 0.252 & \textbf{--} & 0.549 & 0.732 \\
PixelSeg~\cite{zhang2022pixelseg} & 0.243 & \textbf{--} & 0.614 & 0.814 \\
CIMD~\cite{rahman2023ambiguous} & 0.234 & \textbf{--} & 0.587 & \textbf{--} \\
Mose~\cite{gao2023modeling} & 0.234 & \textbf{--} & 0.623 & 0.702 \\
SAMed~\cite{zhang2023customized} & 0.380 & \textbf{--} & 0.357 & 0.703 \\
P$^2$SAM~\cite{huang2024p2sam} & 0.353 & 3.648 & 0.654 & 0.772 \\
\cellcolor{hlgreen}SARFA (ours) & \cellcolor{hlgreen}0.206 & \cellcolor{hlgreen}2.758 & \cellcolor{hlgreen}0.659 & \cellcolor{hlgreen}0.774 \\
\bottomrule
\end{tabular}
}
\label{tab:lidcresults}
\end{table}

\subsection{Implementation Details}

\noindent\textbf{Baselines:} We compared our SARFA against ten state-of-the-art (SOTA) SAM-based and non-SAM baselines: Probabilistic U-Net~\cite{kohl2018probabilistic}, HPU-Net~\cite{kohl2019hierarchical}, PHiSeg~\cite{baumgartner2019phiseg}, SSN~\cite{monteiro2020stochastic}, CAR~\cite{kassapis2021calibrated}, PixelSeg~\cite{zhang2022pixelseg}, CIMD~\cite{rahman2023ambiguous}, Mose~\cite{gao2023modeling}, SAMed~\cite{zhang2023customized}, and $\text{P}^2\text{SAM}$~\cite{huang2024p2sam}. 

\noindent\textbf{Training:} We trained for 10 epochs using a batch size of 1, a learning rate of 0.001, and an image size of 128$\times$128 for the generation of 16 candidate masks. Just 10 epochs were used for training after a convergence analysis (shown in Fig.~\ref{fig:loss_curves}) revealed that both SARFA and our strong baseline P$^2$SAM converged within the first 10 epochs, with no meaningful gains in performance after this point.  For the DPO-specific hyperparameters, we used a $\beta$ of 0.01, a $\lambda$ of 0.05, and a $T$ of 10 to apply DPO every 10 steps. All of these hyperparameters were empirically tuned for optimal model performance. 

\noindent\textbf{Machine Configuration:} The models are trained on a \emph{Intel (R) Xeon (R) w7-2475X, 2600MHz} machine with a dual \emph{NVIDIA A4000X2} GPUs (32GB). 

\begin{figure}[t]
    \centering

    {\small Input \par}
    \vspace{2pt}
    \includegraphics[width = 0.22\columnwidth]{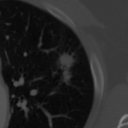}

    \vspace{5pt}

    {\small Ground Truth Masks \par}
    \vspace{2pt}
    \makebox[\columnwidth][c]{
        \includegraphics[width = 0.22\columnwidth]{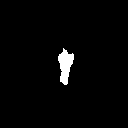}
        \hspace{0.015\columnwidth}
        \includegraphics[width = 0.22\columnwidth]{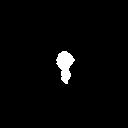}
        \hspace{0.015\columnwidth}
        \includegraphics[width = 0.22\columnwidth]{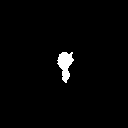}
        \hspace{0.015\columnwidth}
        \includegraphics[width = 0.22\columnwidth]{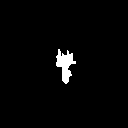}
    }

    \vspace{5pt}

    {\small P$^2$SAM Masks \par}
    \vspace{2pt}
    \makebox[\columnwidth][c]{
        \includegraphics[width = 0.22\columnwidth]{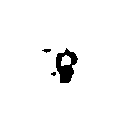}
        \hspace{0.015\columnwidth}
        \includegraphics[width = 0.22\columnwidth]{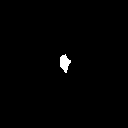}
        \hspace{0.015\columnwidth}
        \includegraphics[width = 0.22\columnwidth]{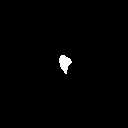}
        \hspace{0.015\columnwidth}
        \includegraphics[width = 0.22\columnwidth]{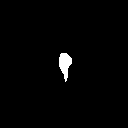}
    }

    \vspace{5pt}

    {\small SARFA Masks \par}
    \vspace{2pt}
    \makebox[\columnwidth][c]{
        \includegraphics[width = 0.22\columnwidth]{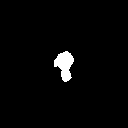}
        \hspace{0.015\columnwidth}
        \includegraphics[width = 0.22\columnwidth]{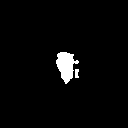}
        \hspace{0.015\columnwidth}
        \includegraphics[width = 0.22\columnwidth]{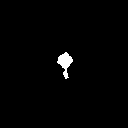}
        \hspace{0.015\columnwidth}
        \includegraphics[width = 0.22\columnwidth]{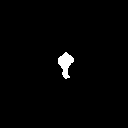}
    }

    \caption{Visual comparison of selected plausible masks generated by both P$^2$SAM and SARFA with the ground truth annotations for the LIDC-IDRI dataset. Note the failed segmentation of P$^2$SAM's first mask compared to the successful segmentation of SARFA's first mask. Additional visualizations are included in the Supplemental Material.}
    \label{fig:qualresults_lidc}
\end{figure}

\noindent\textbf{Evaluation:} For evaluation, we used generalized energy distance (GED), FRD, Hungarian-matched IoU (HM-IoU), and maximum Dice matching ($D_{max}$).

\subsection{Results and Discussion}

\noindent\textbf{Results on LIDC-IDRI:} Table~\ref{tab:lidcresults} reports the results of SARFA compared against ten SOTA ambiguous segmentation models. Comparatively, SARFA performs very well, outperforming each of the ten methods. Compared to the strongest baseline in P$^2$SAM, our SARFA outperforms it marginally on all metrics, with the biggest improvements being noted on the distance-based metrics GED and FRD, indicating that the plausible masks predicted by SARFA are more closely aligned with the ground truth distribution. Fig.~\ref{fig:qualresults_lidc} shows a qualitative comparison of the first four masks generated by both P$^2$SAM and SARFA. As evidenced by the first of P$^2$SAM's predicted masks, SARFA can accurately 
which completely fails to properly segment the lung lesion, SARFA is very capable of outperforming P$^2$SAM in terms of visual output.
 
\begin{figure}[t]
    \centering

    {\small Input \par}
    \vspace{2pt}
    \includegraphics[width = 0.305\columnwidth]{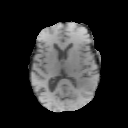}

    \vspace{5pt}

    {\small Ground Truth Masks \par}
    \vspace{2pt}
    \makebox[\columnwidth][c]{
        \includegraphics[width = 0.305\columnwidth]{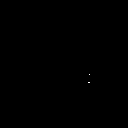}
        \hspace{0.015\columnwidth}
        \includegraphics[width = 0.305\columnwidth]{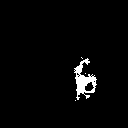}
        \hspace{0.015\columnwidth}
        \includegraphics[width = 0.305\columnwidth]{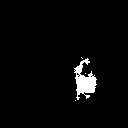}
    }

    \vspace{5pt}

    {\small P$^2$SAM Masks \par}
    \vspace{2pt}
    \makebox[\columnwidth][c]{
        \includegraphics[width = 0.305\columnwidth]{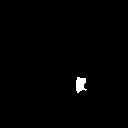}
        \hspace{0.015\columnwidth}
        \includegraphics[width = 0.305\columnwidth]{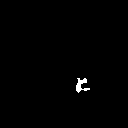}
        \hspace{0.015\columnwidth}
        \includegraphics[width = 0.305\columnwidth]{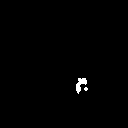}
    }

    \vspace{5pt}

    {\small SARFA Masks \par}
    \vspace{2pt}
    \makebox[\columnwidth][c]{
        \includegraphics[width = 0.305\columnwidth]{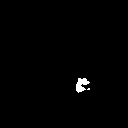}
        \hspace{0.015\columnwidth}
        \includegraphics[width = 0.305\columnwidth]{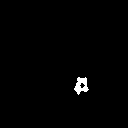}
        \hspace{0.015\columnwidth}
        \includegraphics[width = 0.305\columnwidth]{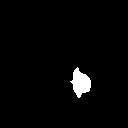}
    }

    \caption{Visual comparison of selected plausible masks generated by both P$^2$SAM and SARFA with the ground truth annotations for the BraTS2017 dataset. Additional visualizations are included in the Supplemental Material.}
    \label{fig:qualresults_brats}
\end{figure}

\noindent\textbf{Results on BraTS2017:} Table~\ref{tab:bratsresults} shows the results of our proposed SARFA against both implementations of the strong baseline P$^2$SAM on the BraTS2017 dataset. Here, SARFA clearly outperforms P$^2$SAM across all metrics, demonstrating the efficacy of SARFA in achieving strong performance across imaging modalities and segmentation tasks. This is further validated by the visual comparison of the segmentation masks produced by SARFA and those produced by P$^2$SAM, as shown in Fig.~\ref{fig:qualresults_brats}.

\noindent\textbf{Hyperparameter Tuning:} In addition to investigating whether FRD is a valid training objective, the findings of which we have already discussed, we have also comprehensively evaluated different configurations of the loss functions used by SARFA. The outcomes of these experiments are shown in Table~\ref{tab:ablation}. To demonstrate that SARFA is capable of performing with varying values for the $K$ candidate mask generation, all experiments in Table~\ref{tab:ablation} were performed using $K = 8$, whereas $K = 16$ was used in the experiments reported in Tables~\ref{tab:lidcresults} and~\ref{tab:bratsresults}. The components evaluated were the configuration of the supervised loss, $\mathcal{L}_\text{sup}$, the temperature of $\mathcal{L}_\text{DPO}$, $\beta$, the weight of $\mathcal{L}_\text{DPO}$, $\lambda$, and the number of steps, $T$, that pass before $\mathcal{L}_\text{DPO}$ is calculated.  We find that the configuration of $\mathcal{L}_\text{sup}$ = $\mathcal{L}_{\text{Dice}}$ + $\mathcal{L}_{\text{CE}}$ + $\mathcal{L}_{\text{Focal}}$, $\beta = 0.1$, $\lambda = 0.05$, and $T = 10$ achieves the best balanced performance across the losses among the examined configurations, thus is the configuration used for our experiments reported in Tables~\ref{tab:lidcresults} and~\ref{tab:bratsresults}.

\begin{table}[t]
\centering
\caption{Comparison of SOTA ambiguous segmentation models on BraTS2017. $\dagger$ indicates that the best model was saved using the lowest validation loss, while $\ddagger$ indicates that the lowest FRD was used to select the best model. The best-performing method across the majority of metrics is highlighted in green.}
\resizebox{\linewidth}{!}{
\begin{tabular}{lcccc}
\toprule
Method & GED($\downarrow$) & FRD($\downarrow$) & HM-IoU($\uparrow$) & $D_{\max}(\uparrow)$ \\
\midrule
P$^2$SAM ($\dagger$) & 8.426 & 9.602 & 0.342 & 0.446 \\
P$^2$SAM ($\ddagger$) & 3.979 & 10.496 & 0.326 & 0.427 \\
\cellcolor{hlgreen}SARFA (ours) & \cellcolor{hlgreen}2.644 & \cellcolor{hlgreen}6.198 & \cellcolor{hlgreen}0.358 & \cellcolor{hlgreen}0.458 \\
\bottomrule
\end{tabular}
}
\label{tab:bratsresults}
\end{table}

\noindent\textbf{Ablation Experiments:} To further assess the contribution of each component in SARFA, we performed two ablation studies. First, we evaluated whether the proposed radiomics-based ranking strategy is more effective for defining the DPO preference pairs than conventional overlap-based alternatives. Table~\ref{tab:dpo_ablation} compares three strategies for selecting the positive and negative masks used in the DPO loss: highest/lowest IoU, highest/lowest Dice score, and lowest/highest FRD. Using FRD to construct the preference pairs achieves the strongest overall performance, with the lowest GED and FRD and the highest HM-IoU and $D_{\max}$ among the evaluated ranking strategies. In particular, FRD-based ranking improves GED from 0.290 to 0.194 compared with Dice-based ranking and from 4.209 to 0.194 compared with IoU-based ranking. It also produces the lowest FRD score, reducing the distance from 3.410 with Dice ranking and 5.513 with IoU ranking to 2.818. These results indicate that selecting preferred masks based on radiomic similarity provides a more effective optimization signal than selecting them using only pixel-level overlap.

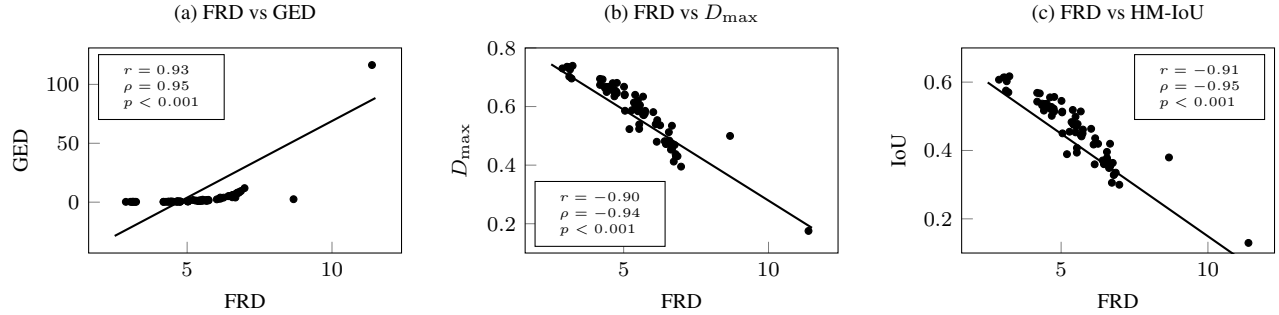
\begin{figure*}[t]
\centering

\resizebox{\linewidth}{!}{

\begin{minipage}[t]{0.33\linewidth}
\centering

\begin{tikzpicture}
\begin{axis}[
    width=\linewidth,
    height=0.75\linewidth,
    xlabel={FRD},
    ylabel={GED},
    title={(a) FRD vs GED},
    xtick pos=bottom,
    ytick pos=left,
    grid=none,
    tick label style={font=\footnotesize},
    label style={font=\footnotesize},
    title style={font=\footnotesize}
]

\addplot[
    only marks,
    mark=*,
    mark size=1.3pt
]
table[
    x=FRD,
    y=GED,
    col sep=comma
]{frd_ged.csv};

\addplot[
    thick,
    domain=2.5:11.5,
    samples=100
]
{13.01*x - 61.2};

\node[
    anchor=north west,
    fill=white,
    draw=black,
    font=\tiny
]
at (rel axis cs:0.03,0.97)
{
\begin{tabular}{l}
$r = 0.93$ \\
$\rho = 0.95$ \\
$p < 0.001$
\end{tabular}
};

\end{axis}
\end{tikzpicture}

\end{minipage}
\hfill

\begin{minipage}[t]{0.33\linewidth}
\centering

\begin{tikzpicture}
\begin{axis}[
    width=\linewidth,
    height=0.75\linewidth,
    xlabel={FRD},
    ylabel={$D_{\max}$},
    title={(b) FRD vs $D_{\max}$},
    ymin=0.1,
    ymax=0.8,
    xtick pos=bottom,
    ytick pos=left,
    grid=none,
    tick label style={font=\footnotesize},
    label style={font=\footnotesize},
    title style={font=\footnotesize}
]

\addplot[
    only marks,
    mark=*,
    mark size=1.3pt
]
table[
    x=FRD,
    y=Dice,
    col sep=comma
]{frd_dice.csv};

\addplot[
    thick,
    domain=2.5:11.5,
    samples=100
]
{-0.062*x + 0.899};

\node[
    anchor=south west,
    fill=white,
    draw=black,
    font=\tiny
]
at (rel axis cs:0.03,0.03)
{
\begin{tabular}{l}
$r = -0.90$ \\
$\rho = -0.94$ \\
$p < 0.001$
\end{tabular}
};

\end{axis}
\end{tikzpicture}

\end{minipage}
\hfill

\begin{minipage}[t]{0.33\linewidth}
\centering

\begin{tikzpicture}
\begin{axis}[
    width=\linewidth,
    height=0.75\linewidth,
    xlabel={FRD},
    ylabel={IoU},
    title={(c) FRD vs HM-IoU},
    ymin=0.1,
    ymax=0.7,
    xtick pos=bottom,
    ytick pos=left,
    grid=none,
    tick label style={font=\footnotesize},
    label style={font=\footnotesize},
    title style={font=\footnotesize}
]

\addplot[
    only marks,
    mark=*,
    mark size=1.3pt
]
table[
    x=FRD,
    y=IoU,
    col sep=comma
]{frd_iou.csv};

\addplot[
    thick,
    domain=2.5:11.5,
    samples=100
]
{-0.060*x + 0.749};

\node[
    anchor=north east,
    fill=white,
    draw=black,
    font=\tiny
]
at (rel axis cs:0.97,0.97)
{
\begin{tabular}{l}
$r = -0.91$ \\
$\rho = -0.95$ \\
$p < 0.001$
\end{tabular}
};

\end{axis}
\end{tikzpicture}

\end{minipage}
}

\caption{
Correlation analysis between FRD and established segmentation quality metrics across all valid candidate masks generated during training ($n = 80$). Each point represents an individual candidate mask and solid lines denote least-squares regression fits. Lower FRD values are associated with lower GED and higher Dice and HM-IoU scores, indicating that radiomic similarity remains strongly aligned with conventional overlap- and distribution-based segmentation metrics.
}
\label{fig:frd_correlation}

\end{figure*}

\begin{table}[t]
\centering
\caption{Segmentation performance under different configurations of the supervised and DPO loss components of SARFA by removing each of our proposed components. The best loss configuration is highlighted in green.}
\resizebox{\linewidth}{!}{
\Large
\begin{tabular}{lccccccc}
\toprule
$\mathcal{L}_{\text{sup}}$ Configuration & $\beta$ & $\lambda$ & $T$ & GED($\downarrow$) & FRD($\downarrow$) & HM-IoU($\uparrow$) & $D_{\max}(\uparrow)$ \\
\midrule
\multirow{4}{*}{$\mathcal{L}_{\text{Dice}} + \mathcal{L}_{\text{CE}} + \mathcal{L}_{\text{Focal}}$}
& \multirow{4}{*}{0.1}
& \multirow{4}{*}{0.05}
& 5  & 0.321 & 3.087 & 0.578 & 0.704 \\
& & & -- & 0.217 & 2.780 & 0.559 & 0.690 \\
& & & \cellcolor{hlgreen}10
  & \cellcolor{hlgreen}0.194
  & \cellcolor{hlgreen}2.818
  & \cellcolor{hlgreen}0.602
  & \cellcolor{hlgreen}0.728 \\
& & & 15 & 6.502 & 5.632 & 0.225 & 0.331 \\
\midrule
 & & -- & & 0.323 & 3.714 & 0.493 & 0.629 \\
$\mathcal{L}_{\text{Dice}}$ + $\mathcal{L}_{\text{CE}}$ + $\mathcal{L}_{\text{Focal}}$ & 0.1 & 0.10 & 10 & 0.252 & 2.903 & 0.600 & 0.722 \\
 & & 0.01 & & 0.341 & 3.275 & 0.579 & 0.706 \\
\midrule
$\mathcal{L}_{\text{Dice}}$ + $\mathcal{L}_{\text{CE}}$ + $\mathcal{L}_{\text{Focal}}$ & 0.2 & 0.05 & 10 & 0.064 & 2.543 & 0.591 & 0.716 \\
\midrule
$\mathcal{L}_{\text{Dice}}$ + $\mathcal{L}_{\text{CE}}$ 
& \multirow{6}{*}{0.1}
& \multirow{6}{*}{0.05}
& \multirow{6}{*}{10}
& 923.046 & 13.133 & 0.008 & 0.015 \\
$\mathcal{L}_{\text{Dice}}$ + $\mathcal{L}_{\text{Focal}}$ & & & & 0.166 & 2.848 & 0.528 & 0.663 \\
$\mathcal{L}_{\text{CE}}$ + $\mathcal{L}_{\text{Focal}}$   & & & & 1.316 & 4.307 & 0.423 & 0.563 \\
$\mathcal{L}_{\text{Focal}}$                        & & & & 0.106 & 3.224 & 0.547 & 0.678 \\
$\mathcal{L}_{\text{CE}}$                           & & & & 1.655 & 5.192 & 0.367 & 0.498 \\
No $\mathcal{L}_\text{sup}$                                           & & & & 38.274 & 8.853 & 0.035 & 0.056 \\
\bottomrule
\end{tabular}}
\label{tab:ablation}
\end{table}

\begin{table}[t]
\centering
\caption{Segmentation performance when the positive and negative pairs for the DPO loss are selected based on the highest and lowest IoU, the highest and lowest Dice score, and the lowest and highest FRD. The best ranking strategy is highlighted in green.}
\resizebox{\linewidth}{!}{
\begin{tabular}{lcccc}
\toprule
Ranking Strategy & GED($\downarrow$) & FRD($\downarrow$) & HM-IoU($\uparrow$) & $D_{\max}(\uparrow)$ \\
\midrule
IoU & 4.209 & 5.513 & 0.301 & 0.402 \\
Dice & 0.290 & 3.410 & 0.543 & 0.673 \\
\cellcolor{hlgreen}FRD & \cellcolor{hlgreen}0.194 & \cellcolor{hlgreen}2.818 & \cellcolor{hlgreen}0.602 & \cellcolor{hlgreen}0.728 \\
\bottomrule
\end{tabular}
}
\label{tab:dpo_ablation}
\end{table}

\begin{table}[t]
\centering
\caption{Ablation experiment on the architectural design of SARFA, where components are removed from the network in the order listed. The best-performing architecture is highlighted in green.}
\resizebox{\linewidth}{!}{
\begin{tabular}{lcccc}
\toprule
Architecture & GED($\downarrow$) & FRD($\downarrow$) & HM-IoU($\uparrow$) & $D_{\max}(\uparrow)$ \\
\midrule
\cellcolor{hlgreen}SARFA & \cellcolor{hlgreen}0.194 & \cellcolor{hlgreen}2.818 & \cellcolor{hlgreen}0.602 & \cellcolor{hlgreen}0.728 \\
– DPO & 5.799 & 5.987 & 0.355 & 0.476 \\
– FRD Ranking & 2.358 & 5.472 & 0.470 & 0.590 \\
– Mask Weighting & 0.557 & 4.593 & 0.527 & 0.656 \\
\bottomrule
\end{tabular}
}
\label{tab:arch_ablation}
\end{table}

Second, we evaluated the architectural contributions of the main SARFA components, as shown in Table~\ref{tab:arch_ablation}. Removing the DPO loss leads to the largest degradation in performance, increasing GED from 0.194 to 5.799 and FRD from 2.818 to 5.987, while reducing HM-IoU from 0.602 to 0.355 and $D_{\max}$ from 0.728 to 0.476. This demonstrates that the DPO objective is essential for learning from ranked candidate masks rather than treating all multimask outputs equally. Removing FRD-based ranking also substantially harms performance, increasing GED to 2.358 and FRD to 5.472, which confirms that the radiomics-driven preference construction is a key component of the proposed framework. Finally, removing the mask weighting mechanism degrades performance across all metrics, though less severely than removing DPO or FRD ranking, with GED increasing to 0.557 and FRD increasing to 4.593. This suggests that mask weighting helps SARFA better combine SAM's candidate outputs, but that its benefit is strongest when paired with radiomics-guided DPO optimization.

\noindent\textbf{Radiomic Alignment as a Segmentation Objective:}
A central hypothesis of this work was that radiomic similarity between predicted and ground-truth regions could serve as a meaningful optimization target for medical image segmentation. While FRD was proposed as a dataset-level evaluation metric, its utility as a training objective had not been explored. Our results provide evidence that radiomic alignment is not only measurable but also strongly associated with conventional indicators of segmentation quality.

Fig.~\ref{fig:frd_correlation} demonstrates a strong relationship between FRD and three established segmentation metrics. Specifically, lower FRD values are associated with lower GED scores and higher HM-IoU and $D_{\max}$ values, with both Pearson ($r$) and Spearman ($\rho$) correlations exceeding 0.90 in magnitude across all comparisons. All observed correlations were determined to be statistically significant ($p < 0.001$), providing strong evidence against the null hypothesis that FRD and the corresponding segmentation metrics are unrelated. This suggests that candidate masks that more closely preserve the radiomic characteristics of the ground-truth region also tend to exhibit stronger spatial agreement with expert annotations. Importantly, this relationship is monotonic as well as linear, indicating that improvements in radiomic similarity consistently correspond to improvements in segmentation quality over a wide range of masks.

\noindent\textbf{Limitations:} Despite the promising results achieved by SARFA, several limitations remain. First, radiomic feature extraction introduces additional computational overhead during training. Large-scale training on high-resolution datasets may require more efficient feature extraction pipelines or approximations of FRD. Second, the current implementation relies on radiomic descriptors extracted using PyRadiomics---alternative representations may capture complementary information and warrant investigation. Another limitation is that the evaluation was conducted on two datasets representing CT and MRI modalities. Furthermore, the current framework constructs preference pairs using only the most preferred and least preferred masks from each candidate set. More sophisticated ranking strategies that exploit the full ordering of candidate masks may provide a richer optimization signal and further improve performance.

\section{Conclusions}
\label{sec:conclusions}

In this paper, we introduced SARFA, a novel probabilistic segmentation model for medical image segmentation that learns during training to produce a distribution of $K$ plausible masks that are radiomically similar to the ground truth distribution. We hypothesized that predicted masks that have similar radiomic features to the ground truth would be better matched anatomically, something that we demonstrated through comprehensive experimentation, demonstrating that the minimization of FRD is a valid training objective for medical segmentation models. Additionally, we demonstrate that incorporating a loss function based on DPO in combination with a standard supervised loss helps learn this training objective. Future work will consist of scaling this architecture to different imaging modalities and segmentation tasks, as well as gauging its performance when additional multimodal information, such as patient metadata, is provided to the model during training.

{
    \small
    \bibliographystyle{ieeenat_fullname}
    \bibliography{main}
}

\appendix
\onecolumn

\section{SARFA Training Algorithm}

A formal algorithm showing the training procedure of SARFA is shown in Algorithm~\ref{alg:sarfa}. 

\begin{algorithm*}[]
\caption{SARFA Training Procedure}
\label{alg:sarfa}
\begin{algorithmic}[1]
\Require Dataset $\mathcal{D}$, SAM checkpoint $C$, epochs $E$, learning rate $\eta$, DPO interval $K$, DPO weight $\lambda$
\Ensure Trained policy model $\pi_\theta$

\State $\pi_\theta \gets$ initialize LoRA-SAM from checkpoint $C$
\State $\pi_{\mathrm{ref}} \gets$ frozen copy of $\pi_\theta$
\State $\mathbf{w} \gets$ initialize learnable mask weights
\State $\phi \gets$ initialize radiomics extractor
\State $(\mu, \sigma) \gets$ radiomics statistics from ground-truth masks in $\mathcal{D}$

\For{$e = 1$ to $E$}
    \State set $\pi_\theta$ and $\mathbf{w}$ to train mode

    \For{each mini-batch $(x, y) \in \mathcal{D}$}
        \State $(M, s_\theta) \gets \pi_\theta(x)$ \Comment{$M = \{\hat{m}_1, \ldots, \hat{m}_8\}$}
        \State $\hat{m}_{w} \gets$ weighted combination of $M$ using $\mathbf{w}$
        \State $L_{\mathrm{sup}} \gets$ segmentation loss using $M$, $\hat{m}_{w}$, and $y$
        \State $L_{\mathrm{DPO}} \gets 0$

        \If{current step is divisible by $K$}
            \State $z_{\mathrm{gt}} \gets$ normalized radiomics features of $(x, y)$

            \For{each candidate mask $\hat{m}_j \in M$}
                \State $z_j \gets$ normalized radiomics features of $(x, \hat{m}_j)$
                \State $d_j \gets$ distance between $z_j$ and $z_{\mathrm{gt}}$
            \EndFor

            \State $c \gets$ candidate with smallest $d_j$
            \State $r \gets$ candidate with largest $d_j$

            \If{$c \neq r$}
                \State $s_{\mathrm{ref}} \gets \pi_{\mathrm{ref}}(x)$
                \State $\Delta_\theta \gets$ policy score margin between $c$ and $r$ using $s_\theta$
                \State $\Delta_{\mathrm{ref}} \gets$ reference score margin between $c$ and $r$ using $s_{\mathrm{ref}}$
                \State $L_{\mathrm{DPO}} \gets$ DPO loss from $\Delta_\theta$, $\Delta_{\mathrm{ref}}$
            \EndIf
        \EndIf

        \State $L \gets L_{\mathrm{sup}} + \lambda L_{\mathrm{DPO}}$
        \State update $\pi_\theta$ and $\mathbf{w}$ using $L$
    \EndFor

    \State $\mathrm{FRD} \gets$ evaluate radiomics distance on epoch predictions
    \If{$\mathrm{FRD}$ improves}
        \State save checkpoint
    \EndIf
\EndFor

\State \Return $\pi_\theta$

\end{algorithmic}
\end{algorithm*}

\clearpage

\section{Additional Visualizations}
Figs.~\ref{fig:lidc1} and~\ref{fig:brats1} show extended visualizations of the samples used for Figs.~\ref{fig:qualresults_lidc} and~\ref{fig:qualresults_brats} of the main text up to eight output masks. Note that while both P$^2$SAM and SARFA are probabilistic methods that can produce multiple mask variants, P$^2$SAM occasionally fails to segment samples (instead segmenting the background), something that is not observed in our SARFA visualizations. Visualizations for additional samples of the LIDC-IDRI and BraTS datasets are shown in Figs.~\ref{fig:lidc2} and~\ref{fig:brats2}.

\begin{center}
\begin{minipage}{\textwidth}
    \centering

    {\small Input \par}
    \vspace{2pt}
    \includegraphics[width = 0.11\textwidth]{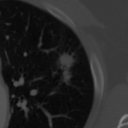}

    \vspace{5pt}

    {\small Ground Truth Masks \par}
    \vspace{2pt}
    \makebox[\textwidth][c]{%
        \includegraphics[width = 0.11\textwidth]{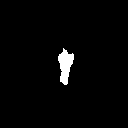}
        \hspace{0.01\textwidth}
        \includegraphics[width = 0.11\textwidth]{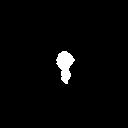}
        \hspace{0.01\textwidth}
        \includegraphics[width = 0.11\textwidth]{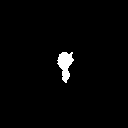}
        \hspace{0.01\textwidth}
        \includegraphics[width = 0.11\textwidth]{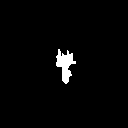}
    }

    \vspace{5pt}

    {\small P$^2$SAM Masks \par}
    \vspace{2pt}
    \makebox[\textwidth][c]{%
        \includegraphics[width = 0.11\textwidth]{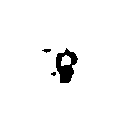}
        \hspace{0.01\textwidth}
        \includegraphics[width = 0.11\textwidth]{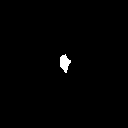}
        \hspace{0.01\textwidth}
        \includegraphics[width = 0.11\textwidth]{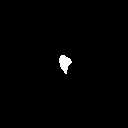}
        \hspace{0.01\textwidth}
        \includegraphics[width = 0.11\textwidth]{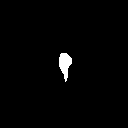}
        \hspace{0.01\textwidth}
        \includegraphics[width = 0.11\textwidth]{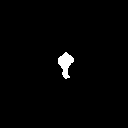}
        \hspace{0.01\textwidth}
        \includegraphics[width = 0.11\textwidth]{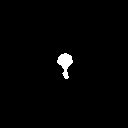}
        \hspace{0.01\textwidth}
        \includegraphics[width = 0.11\textwidth]{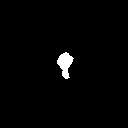}
        \hspace{0.01\textwidth}
        \includegraphics[width = 0.11\textwidth]{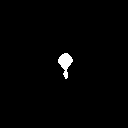}
    }

    \vspace{5pt}

    {\small SARFA Masks \par}
    \vspace{2pt}
    \makebox[\textwidth][c]{%
        \includegraphics[width = 0.11\textwidth]{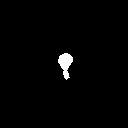}
        \hspace{0.01\textwidth}
        \includegraphics[width = 0.11\textwidth]{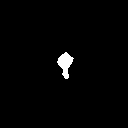}
        \hspace{0.01\textwidth}
        \includegraphics[width = 0.11\textwidth]{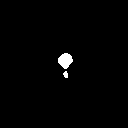}
        \hspace{0.01\textwidth}
        \includegraphics[width = 0.11\textwidth]{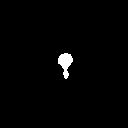}
        \hspace{0.01\textwidth}
        \includegraphics[width = 0.11\textwidth]{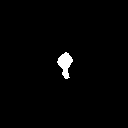}
        \hspace{0.01\textwidth}
        \includegraphics[width = 0.11\textwidth]{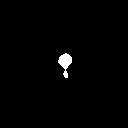}
        \hspace{0.01\textwidth}
        \includegraphics[width = 0.11\textwidth]{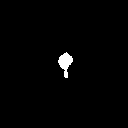}
        \hspace{0.01\textwidth}
        \includegraphics[width = 0.11\textwidth]{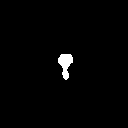}
    }

    \captionof{figure}{Visual comparison of plausible masks generated by both P$^2$SAM and SARFA with the ground truth annotations for the LIDC-IDRI dataset. Note the failed segmentation of P$^2$SAM's first mask compared to the successful segmentation of SARFA's first mask.}
    \label{fig:lidc1}
\end{minipage}
\end{center}

\clearpage

\begin{center}
\begin{minipage}{\textwidth}
    \centering

    {\small Input \par}
    \vspace{2pt}
    \includegraphics[width = 0.11\textwidth]{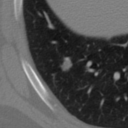}

    \vspace{5pt}

    {\small Ground Truth Masks \par}
    \vspace{2pt}
    \makebox[\textwidth][c]{%
        \includegraphics[width = 0.11\textwidth]{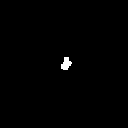}
        \hspace{0.01\textwidth}
        \includegraphics[width = 0.11\textwidth]{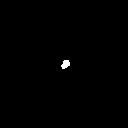}
        \hspace{0.01\textwidth}
        \includegraphics[width = 0.11\textwidth]{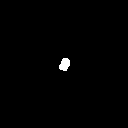}
        \hspace{0.01\textwidth}
        \includegraphics[width = 0.11\textwidth]{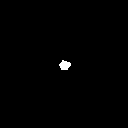}
    }

    \vspace{5pt}

    {\small P$^2$SAM Masks \par}
    \vspace{2pt}
    \makebox[\textwidth][c]{%
        \includegraphics[width = 0.11\textwidth]{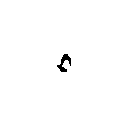}
        \hspace{0.01\textwidth}
        \includegraphics[width = 0.11\textwidth]{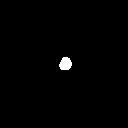}
        \hspace{0.01\textwidth}
        \includegraphics[width = 0.11\textwidth]{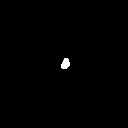}
        \hspace{0.01\textwidth}
        \includegraphics[width = 0.11\textwidth]{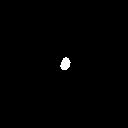}
        \hspace{0.01\textwidth}
        \includegraphics[width = 0.11\textwidth]{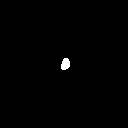}
        \hspace{0.01\textwidth}
        \includegraphics[width = 0.11\textwidth]{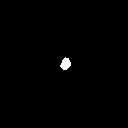}
        \hspace{0.01\textwidth}
        \includegraphics[width = 0.11\textwidth]{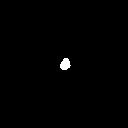}
        \hspace{0.01\textwidth}
        \includegraphics[width = 0.11\textwidth]{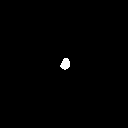}
    }

    \vspace{5pt}

    {\small SARFA Masks \par}
    \vspace{2pt}
    \makebox[\textwidth][c]{%
        \includegraphics[width = 0.11\textwidth]{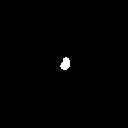}
        \hspace{0.01\textwidth}
        \includegraphics[width = 0.11\textwidth]{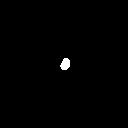}
        \hspace{0.01\textwidth}
        \includegraphics[width = 0.11\textwidth]{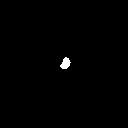}
        \hspace{0.01\textwidth}
        \includegraphics[width = 0.11\textwidth]{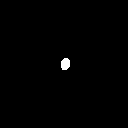}
        \hspace{0.01\textwidth}
        \includegraphics[width = 0.11\textwidth]{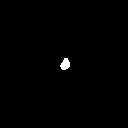}
        \hspace{0.01\textwidth}
        \includegraphics[width = 0.11\textwidth]{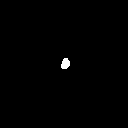}
        \hspace{0.01\textwidth}
        \includegraphics[width = 0.11\textwidth]{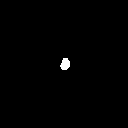}
        \hspace{0.01\textwidth}
        \includegraphics[width = 0.11\textwidth]{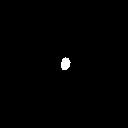}
    }

    \captionof{figure}{Additional visualizations of plausible masks generated by both P$^2$SAM and SARFA with the ground truth annotations for the LIDC-IDRI dataset. Note the failed segmentation of P$^2$SAM's first mask compared to the successful segmentation of SARFA's first mask.}
    \label{fig:lidc2}
\end{minipage}
\end{center}

\clearpage

\begin{center}
\begin{minipage}{\textwidth}
    \centering

    {\small Input \par}
    \vspace{2pt}
    \includegraphics[width = 0.11\textwidth]{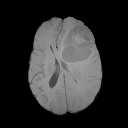}

    \vspace{5pt}

    {\small Ground Truth Masks \par}
    \vspace{2pt}
    \makebox[\textwidth][c]{%
        \includegraphics[width = 0.11\textwidth]{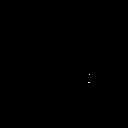}
        \hspace{0.01\textwidth}
        \includegraphics[width = 0.11\textwidth]{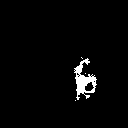}
        \hspace{0.01\textwidth}
        \includegraphics[width = 0.11\textwidth]{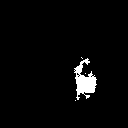}
    }

    \vspace{5pt}

    {\small P$^2$SAM Masks \par}
    \vspace{2pt}
    \makebox[\textwidth][c]{%
        \includegraphics[width = 0.11\textwidth]{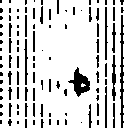}
        \hspace{0.01\textwidth}
        \includegraphics[width = 0.11\textwidth]{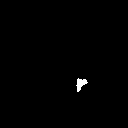}
        \hspace{0.01\textwidth}
        \includegraphics[width = 0.11\textwidth]{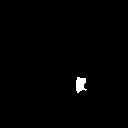}
        \hspace{0.01\textwidth}
        \includegraphics[width = 0.11\textwidth]{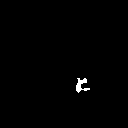}
        \hspace{0.01\textwidth}
        \includegraphics[width = 0.11\textwidth]{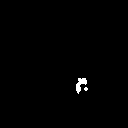}
        \hspace{0.01\textwidth}
        \includegraphics[width = 0.11\textwidth]{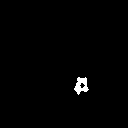}
        \hspace{0.01\textwidth}
        \includegraphics[width = 0.11\textwidth]{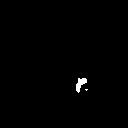}
        \hspace{0.01\textwidth}
        \includegraphics[width = 0.11\textwidth]{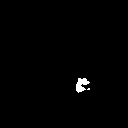}
    }

    \vspace{5pt}

    {\small SARFA Masks \par}
    \vspace{2pt}
    \makebox[\textwidth][c]{%
        \includegraphics[width = 0.11\textwidth]{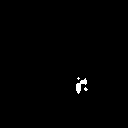}
        \hspace{0.01\textwidth}
        \includegraphics[width = 0.11\textwidth]{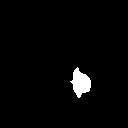}
        \hspace{0.01\textwidth}
        \includegraphics[width = 0.11\textwidth]{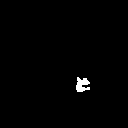}
        \hspace{0.01\textwidth}
        \includegraphics[width = 0.11\textwidth]{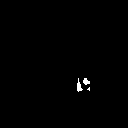}
        \hspace{0.01\textwidth}
        \includegraphics[width = 0.11\textwidth]{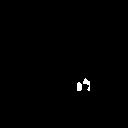}
        \hspace{0.01\textwidth}
        \includegraphics[width = 0.11\textwidth]{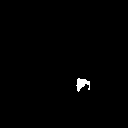}
        \hspace{0.01\textwidth}
        \includegraphics[width = 0.11\textwidth]{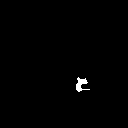}
        \hspace{0.01\textwidth}
        \includegraphics[width = 0.11\textwidth]{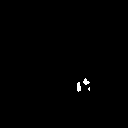}
    }

    \captionof{figure}{Visual comparison of plausible masks generated by both P$^2$SAM and SARFA with the ground truth annotations for the BraTS2017 dataset.}
    \label{fig:brats1}
\end{minipage}
\end{center}

\clearpage

\begin{center}
\begin{minipage}{\textwidth}
    \centering

    {\small Input \par}
    \vspace{2pt}
    \includegraphics[width = 0.11\textwidth]{supplemental_images/brats/image.png}

    \vspace{5pt}

    {\small Ground Truth Masks \par}
    \vspace{2pt}
    \makebox[\textwidth][c]{%
        \includegraphics[width = 0.11\textwidth]{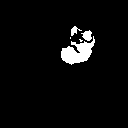}
        \hspace{0.01\textwidth}
        \includegraphics[width = 0.11\textwidth]{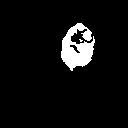}
        \hspace{0.01\textwidth}
        \includegraphics[width = 0.11\textwidth]{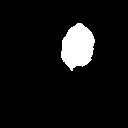}
    }

    \vspace{5pt}

    {\small P$^2$SAM Masks \par}
    \vspace{2pt}
    \makebox[\textwidth][c]{%
        \includegraphics[width = 0.11\textwidth]{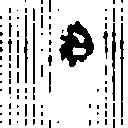}
        \hspace{0.01\textwidth}
        \includegraphics[width = 0.11\textwidth]{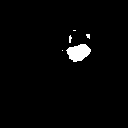}
        \hspace{0.01\textwidth}
        \includegraphics[width = 0.11\textwidth]{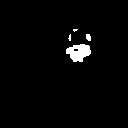}
        \hspace{0.01\textwidth}
        \includegraphics[width = 0.11\textwidth]{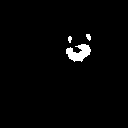}
        \hspace{0.01\textwidth}
        \includegraphics[width = 0.11\textwidth]{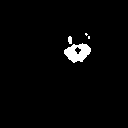}
        \hspace{0.01\textwidth}
        \includegraphics[width = 0.11\textwidth]{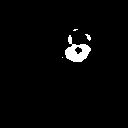}
        \hspace{0.01\textwidth}
        \includegraphics[width = 0.11\textwidth]{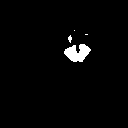}
        \hspace{0.01\textwidth}
        \includegraphics[width = 0.11\textwidth]{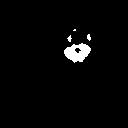}
    }

    \vspace{5pt}

    {\small SARFA Masks \par}
    \vspace{2pt}
    \makebox[\textwidth][c]{%
        \includegraphics[width = 0.11\textwidth]{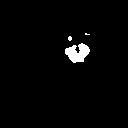}
        \hspace{0.01\textwidth}
        \includegraphics[width = 0.11\textwidth]{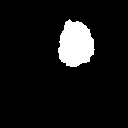}
        \hspace{0.01\textwidth}
        \includegraphics[width = 0.11\textwidth]{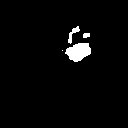}
        \hspace{0.01\textwidth}
        \includegraphics[width = 0.11\textwidth]{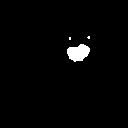}
        \hspace{0.01\textwidth}
        \includegraphics[width = 0.11\textwidth]{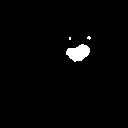}
        \hspace{0.01\textwidth}
        \includegraphics[width = 0.11\textwidth]{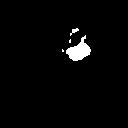}
        \hspace{0.01\textwidth}
        \includegraphics[width = 0.11\textwidth]{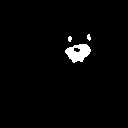}
        \hspace{0.01\textwidth}
        \includegraphics[width = 0.11\textwidth]{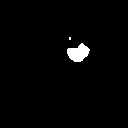}
    }

    \captionof{figure}{Additional visual comparisons of plausible masks generated by both P$^2$SAM and SARFA with the ground truth annotations for the BraTS2017 dataset.}
    \label{fig:brats2}
\end{minipage}
\end{center}

\end{document}